\documentclass[conference]{IEEEtran}
\usepackage{geometry}
\usepackage{graphicx}
\usepackage[backend = bibtex,
			style = numeric,
			sorting = none,
			]{biblatex}
			
{\footnotesize
\bibliography{references.bib}}
\usepackage{xcolor}
\usepackage{comment}
\usepackage{varwidth}
\usepackage{amsmath}
\usepackage{enumitem}   
\usepackage{dblfloatfix} 
\usepackage{booktabs}
\usepackage{caption}
\usepackage{multirow}
\usepackage{subfigure} 
\usepackage{amsmath}
\usepackage{amsfonts}
\usepackage{amssymb}
\usepackage[ruled,vlined]{algorithm2e}
\usepackage{amsbsy}
\usepackage{pifont}
\usepackage{todonotes}
\usepackage{array}
\newcommand{\xmark}{\ding{53}}
\usepackage{balance} % In case references run across two columns on a final page, keep them at an even height in each column.
\usepackage{multirow}

\newcommand{\eg}[1]{}
\renewcommand{\eg}[1]{e.g. {#1}}

\newcommand{\ie}[1]{}
\renewcommand{\ie}[1]{i.e. {#1}}

\newcommand*{\DEBUG}{}
\ifdefined\DEBUG
\newcommand{\anote}[1]{\textcolor{purple}{[AG:#1]}}
\newcommand{\gnote}[1]{\textcolor{magenta}{[GKT:#1]}}
\newcommand{\jnote}[1]{\textcolor{olive}{[JCD:#1]}}
\newcommand{\ynote}[1]{\textcolor{red}{[YHL:#1]}}
\newcommand{\cnote}[1]{\textcolor{blue}{[CT:#1]}}
\newcommand{\xnote}[1]{\textcolor{royal}{[XH:#1]}}
\newcommand{\hnote}[1]{\textcolor{green}{[HW:#1]}}
\fi

\ifdefined\NODEBUG
\newcommand{\anote}[1]{}
\newcommand{\gnote}[1]{}
\newcommand{\jnote}[1]{}
\newcommand{\ynote}[1]{}
\newcommand{\cnote}[1]{}
\newcommand{\xnote}[1]{}
\newcommand{\hnote}[1]{}
\newcommand{\nnote}[1]{}
\newcommand{\delete}[1]{}
\fi

\begin{document}

\title{Efficient Computer Vision on Edge Devices with Pipeline-Parallel Hierarchical Neural Networks}

\author{\IEEEauthorblockN{Abhinav Goel,
Caleb Tung, Xiao Hu,
George K. Thiruvathukal\IEEEauthorrefmark{1}, James C. Davis, Yung-Hsiang Lu}
\IEEEauthorblockA{Purdue University, School of Electrical and Computer Engineering, West Lafayette, IN, USA \\
\IEEEauthorrefmark{1}Loyola University Chicago, Department of Computer Science, Chicago, IL, USA}}

\maketitle

\makeatletter
\def\ps@IEEEtitlepagestyle{%
  \def\@oddfoot{\mycopyrightnotice}%
  \def\@evenfoot{}%
}
\makeatother
\def\mycopyrightnotice{%
  \begin{minipage}{\textwidth}
    \footnotesize
    ~ \hfill\\~\\
  \end{minipage}
  \gdef\mycopyrightnotice{}% just in case
}

\begin{abstract}
% Running Computer Vision (CV) on edge devices has many applications, \eg emergency response and tracking.
% However, most CV algorithms, such as Deep Neural Networks (DNNs), are too computationally expensive for inference on resource-constrained edge devices. 
% To run DNNs for edge-applications, the existing work either offloads DNN computing to the cloud or increases the efficiency of DNNs via model compression. These techniques overlook the potential of parallel DNN inference over distributed edge devices.
% %Hierarchical DNNs are one such technique. Here, a large DNN is broken into a hierarchy of small DNNs to reduce redundant computation by performing intermediate classifications at every level of the hierarchy.
% In this work we consider the hierarchical DNN architecture, where a single large DNN is split into a hierarchy of small DNNs to increase efficiency on edge devices.
% We design novel methods to distribute the inference of hierarchical DNNs across multiple edge devices. 
% The proposed method balances loads across the collaborating edge devices and creates an inference pipeline to facilitate the processing of multiple 
% %for varying image sizes, DNN sizes, and input sequence lengths to identify the scenarios in which the proposed pipelined distribution is desirable. We also compare our pipelined distribution approach with the straightforward round-robin distribution of images.
% We observe that on average our approach achieves X\% higher throughput and Y\% less energy consumption per device per image, with only a Z\% increase in latency.

Computer vision on low-power edge devices enables applications including search-and-rescue and security.
State-of-the-art computer vision algorithms, such as Deep Neural Networks (DNNs), are too large for inference on low-power edge devices.
To improve efficiency, some existing approaches parallelize DNN inference across multiple edge devices.
However, these techniques introduce significant communication and synchronization overheads or are unable to balance workloads across devices. 
This paper demonstrates that the 
hierarchical DNN architecture is well suited for parallel processing on multiple edge devices.
We design a novel method that creates a parallel inference pipeline for computer vision problems that use hierarchical DNNs. 
The method balances loads across the collaborating devices and reduces communication costs to facilitate the processing of multiple video frames simultaneously with higher throughput. Our
experiments consider a representative computer vision problem where image recognition is performed on each video frame, running on multiple Raspberry Pi 4Bs.
With four collaborating low-power edge devices, our approach achieves 3.21$\times$ higher throughput, 68\% less energy consumption per device per frame, and a 58\% decrease in memory when compared with existing single-device hierarchical DNNs. 
\end{abstract}
%for varying image sizes, DNN sizes, and input sequence lengths to identify the scenarios in which the proposed pipelined distribution is desirable. We also compare our pipelined distribution approach with the straightforward round-robin distribution of images.

\begin{IEEEkeywords}
Parallel edge computing, hierarchical DNNs.
\end{IEEEkeywords}

\section{Introduction}

Deep Neural Networks (DNNs) are the state-of-the-art techniques to perform computer vision tasks on video streams. 
Because of the significant energy and computation resource requirements of DNNs, video stream processing is usually performed on the Cloud~\cite{lpirc}.
However, applications with strict throughput, privacy, or network bandwidth constraints must be handled locally~\cite{FALCON}.
Increasing the efficiency of DNNs will enable more low-power edge devices to process visual data without offloading.
%enable low-power edge devices  to perform tasks like traffic monitoring and surveillance~\cite{8050296}..
%This limitation needs to be overcome for computer vision to be deployed on resource-constrained edge devices.  

%Multiple techniques (e.g., quantization~\cite{Han2015} and compression~\cite{Mob}) have been developed to increase DNN efficiency. These techniques, however, do not account for the fact that multiple edge devices are often deployed on the same network~\cite{mednn, modnn}.  Distributed edge computing systems offer several important advantages, including more better utilization of local computing resources, more privacy, and less dependency on network bandwidth, etc.

%Often multiple devices are deployed together for tasks like traffic monitoring and surveillance~\cite{sara}.

%\jnote{The flow in this paragraph (first two sentences) is still a little bit off. Can you try a version where the second half of this paragraph is merged with the first paragraph, and then the rest of the second paragraph is merged with (what is now the) third paragraph to present ``prior work towards this application context''?}

Existing efforts to increase DNN efficiency are largely focused on single-device inference~\cite{goel2020survey, Mob}.
However, low-power edge devices are commonly deployed in a network, \eg to enable monitoring of multiple angles of a traffic intersection or a construction site~\cite{sara}.
If these networks have spare computing resources, then \textit{parallel} inference would allow the devices to share resources for faster data processing~\cite{iot1, modnn}. For example, if an edge device is not powerful enough to provide the required response time, the device could partition the DNN, and transmit the partitioned tasks to other devices~\cite{pipeline}. 
%at a given time, portions of a construction site may not have any workers; the edge devices overlooking these portions have spare resources (only see the background) to share with the other devices. 
%\todo[inline]{This seems a contrived example. Can you cite some references or give more examples?}
%In most cases, we expect that edge device networks have spare computing resources because they are driven by unpredictable and/or imbalanced human behavior~\cite{iot1}.
%
% Parallel DNN inference allows devices to 
%\jnote{I'm not sure you need to make this argument here. It should certainly go in ``system design'', but here you might say just ``If these networks are overprovisioned, then parallel inference would allow the device network to...''}
%\jnote{add a word like ``singly'' to indicate that the issue is using ONE device. See next comment.}
%\jnote{The next sentence begins with the wrong clause, so the argument is disjointed. Try a sentence like ``However, low-power edge devices are commonly deployed in a network, \eg to enable monitoring of multiple angles of a traffic intersection or a construction site.'' You can use that to launch the rest of the paragraph.}
%These existing techniques focus on single-device inference, and do not consider the scenarios where networks of multiple edge devices are used together \eg traffic intersections and construction sites.

\begin{figure}[b!]
        \vspace{-0.15in}
        \centering
        \subfigure[]{\includegraphics[height = 1.37in]{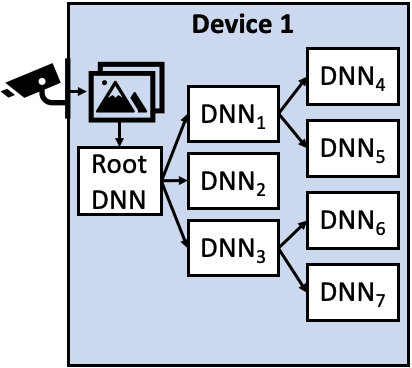}
        }
        \subfigure[]{\includegraphics[height = 1.37in]{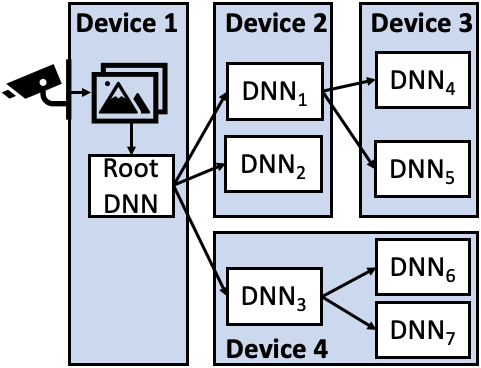}}
        \caption{(a)~Existing hierarchical DNNs: multiple small DNNs in the form of a tree. All DNNs along a single root-leaf path process a video frame before the next frame can be processed~\cite{todaes, treecnn, islped}. (b)~Our method: after the root DNN at device 1 processes a frame, the frame is passed onto another device. This allows device 1 to start processing the next frame and creates an inference pipeline to increase throughput.
        }
        \label{fig:pipeline}

\end{figure}

%\jnote{Topic sentence does not include ``with tradeoffs'', yet you spend half the paragraph pointing out tradeoffs. There is no criticism of data parallelism like there is of model and pipeline parallelism. I suggest you try another pass, with the THIRD sentence reading as: ``data parallelism does X; model...Y, pipeline...Z. and then a FOURTH-FIFTH sentence stating that pipeline seems most promising for the target workload. That leads us nicely into the NEXT paragraph where you can say the shortcoming of the SOTA and your approach. HOWEVER, you introduce video stream processing HERE as your target workload. This is the first time you have done so. It should be earlier, no? Perhaps in the Title, and in paragraph 1?}

Some existing works perform parallel DNN inference on edge devices.
These methods can be classified as (a)~Data parallelism: each collaborating edge device processes a subset of the inputs with the assumption that each device can run the entire DNN~\cite{modnn}; (b)~Model parallelism: each DNN layer is partitioned across multiple devices, but requires extensive inter-device communication to map inputs and reduce outputs~\cite{strads}; (c)~Pipeline parallelism: the DNN is partitioned into sets of consecutive layers.
Each set is run on a different device; after one device processes a frame, the frame is passed onto another device. This allows the first device to process the next frame for improved throughput~\cite{pipeline2}. Pipeline parallelism is most suitable for improving the throughput of video stream processing but is currently limited because conventional DNNs have a large variance in resource requirements across layers~\cite{mednn}.

% \jnote{I am not sure this detail is required here. The previous sentence is fine. You should the note ``up to 15x'', perhaps in parentheses. The detail should go into System Design where you talk in more detail about resource variance.}
% On a Raspberry Pi 3, the inference time of largest fully-connected layer of VGG-16 is $\sim 15\times$ larger than the inference time of the smallest convolution layer~\cite{pipeline}.

%By collaborating and sharing computing resources between nearby devices, DNNs can run faster and more efficiently on the edge. 

%The sharing of resources among nearby edge devices is useful because edge-based workloads are often temporal~\cite{temporalpapers}. For example, as seen in Fig.~\ref{fig:motivation}, when one camera is monitoring moving traffic and is processing multiple frames per second, nearby cameras may be overlooking stationary vehicles and processing just a few frames per second. The busy devices can offload some of their computing to nearby idle devices to reduce frame misses. 

We observe that the recent hierarchical DNN architecture~\cite{treecnn, todaes} is well suited for pipeline parallelism.
This architecture is depicted in Fig.~\ref{fig:pipeline}(a) and detailed in Section~\ref{sec:back}. 
The small DNNs of the hierarchy can be partitioned to run independently on collaborating devices without a large cross-device resource variance, as seen in Fig.~\ref{fig:pipeline}(b).
Using this insight, this paper proposes a novel technique to perform pipeline-parallel inference of hierarchical DNNs. Our method partitions hierarchical DNNs in a way that balances workloads and reduces communication costs. We show that, when partitioning the hierarchy, it is advantageous to consider the hierarchy structure and the processing time of each DNN.

To evaluate this approach, our experiments compare the video stream processing performance of the proposed method with state-of-the-art techniques~\cite{Mob, modnn, pipeline, pipeline2, todaes} in terms of frames per second (FPS), latency, memory and numbers of operations, and energy consumption per device.
We vary the hierarchical DNN structures, input resolutions, video lengths, and the number of devices to show that the proposed technique improves throughput for different types of workloads. These experiments are performed using standard computer vision  datasets. 
We observe a 3.21$\times$ increase in FPS, and 60\%, 58\%, 68\% decrease in operations, memory, and energy requirements, respectively. These gains are achieved when using four collaborating Raspberry Pi 4Bs connected via Ethernet.

\section{Background and Related Work}

%to reduce the redundant computation. We do not use single-stage detectors because they use very large DNNs, from which redundancies can not be removed easily. 

%These methods first detect all the objects in an image and then count the number of instances of the queried object. Most object detectors are not suitable for low-power devices because they require large, power-hungry DNNs to determine the location and size of every object in the image. YOLO~\cite{Yolo9000}, Single Shot Detector (SSD)~\cite{SSD}, and SqueezeDet~\cite{squeezedet_2019} are single-stage object detectors. They are faster but less accurate than two-stage detectors like Faster RCNN (RPN + Classifier)~\cite{FRCNN}. 

\label{sec:back}

\subsection{Hierarchical Deep Neural Networks}

Hierarchical DNNs use multiple small DNNs in the form of a tree, as seen in Fig.~\ref{fig:pipeline}(a)~\cite{todaes, treecnn, islped, 9502480}. Each small DNN specializes in an intermediate classification between groups of similar categories. 
In each level of the hierarchy, a small DNN uses the activation map of its parent and makes an intermediate classification into progressively smaller groups, until a leaf DNN provides the final output (\eg DNN\textsubscript{4} in Fig.~\ref{fig:pipeline}). Existing techniques consider the training~\cite{treecnn}, design~\cite{FALCON, todaes}, or applications~\cite{islped} of hierarchical DNNs on one device.

%\jnote{Review this change --- new topic sentence and re-org'd a bit. Is it clearer?}
Hierarchical DNNs offer an energy-accuracy tradeoff.
Since each input is only processed by the small DNNs along one path from the root to a leaf, hierarchical DNNs perform inference more efficiently than conventional DNNs~\cite{FALCON, todaes}, 
reducing energy consumption by $\sim$50\%.
However, they also decrease accuracy by $\sim$4\%, because errors propagate from parent to child DNN.
As this tradeoff may be acceptable in practice~\cite{todaes}, we investigate improving hierarchical DNN processing throughput by performing pipeline-parallel inference.

\begin{table}[b!]
\vspace{-0.1in}
\caption{Comparison of the proposed method with existing methods.
    H-DNN: Hierarchical DNN.}
\centering
\begin{tabular}{lcccccl}\toprule
           \multicolumn{1}{c}{\begin{tabular}[c]{@{}c@{}}Technique\end{tabular}} & \begin{tabular}[c]{@{}c@{}}H-DNN\end{tabular} & \multicolumn{1}{c}{\begin{tabular}[c]{@{}c@{}}Parallelism\\ Type\end{tabular}} & \begin{tabular}[c]{@{}c@{}}Load\\ Balance\end{tabular} & \begin{tabular}[c]{@{}c@{}}Comm.\\ Efficient\end{tabular} \\\midrule
Howard et al.~\cite{Mob}       & \xmark & None & - & - \\
Mao et al.~\cite{modnn}           & \xmark &    Data   & $\checkmark$ & \xmark \\
Zhang et al.~\cite{pipeline}      & \xmark &   Pipeline   & \xmark & $\checkmark$ \\
Hadidi et al.~\cite{pipeline2}   & \xmark & Pipeline + Model     & $\checkmark$ & \xmark\\\midrule
Goel et al.~\cite{todaes}        & $\checkmark$  &  None    & - & -  \\\midrule
Our Method & $\checkmark$ &  Pipeline    & $\checkmark$ & $\checkmark$ \\\bottomrule
\end{tabular}

\label{tab:contributions}
\end{table}

% \textbf{Reducing Deep Neural Network Memory:}
% Marculescu et al.~\cite{iccad} survey DNN designs and discuss the need to reduce memory for efficient inference. Quantization~\cite{flightnn} and pruning methods~\cite{Han2015} are used to reduce memory. MobileNet~\cite{Mob} uses a layer compression technique to reduce the number of parameters and operations.
% These methods focus on efficient inference on a single device. Our proposed method focuses on parallel and low-power inference of hierarchical DNNs.

% \begin{figure}[t!]
%         \centering
%         \includegraphics[width = 0.5\textwidth]{Figures/caltech.png}
%         \caption{
%         Subset of the hierarchical DNN constructed for the CALTECH-256 dataset using the method of Goel et al.~\cite{todaes}.
%         To perform parallel inference across multiple devices, a hierarchical DNN must be partitioned.
%         %TO RELATED WORK!
%         %\jnote{I moved it part of the way, but needs to be referenced in II still. HOWEVER, note that in II.a you reference Figure 1(a), so it's not clear why we need this figure here. I think it feels redundant relative to Figure 1 --- a verbal description of how a hierarchy can be formed, \eg on domain semantics like animals vs. vehicles, seems intuitive without paying the space for this figure. And then in III intro you also reference Figure 1. I'm just not convinced you are getting much value from this figure. This is not a paper about H-DNNS...}
%         }
%         \label{fig:caltech}

% \end{figure}

\subsection{Three Categories of Parallel Edge Computing}

\textit{Data parallelism:} The input data is partitioned and processed independently by the collaborating devices. Splitting a single input frame impacts the spatial locality of objects and lowers DNN accuracy~\cite{wang2021sensAI}. Most data-parallel techniques distribute activation maps to perform convolution operations in parallel~\cite{mednn, modnn}. MoDNN~\cite{modnn} %DeepThings~\cite{deepthings} 
divides each activation map into overlapping grid cells.
MeDNN~\cite{mednn} distributes the activation channels (instead of grids) across devices to avoid repeated operations. All data parallelism techniques assume that each edge device has the capacity to run the entire DNN~\cite{cmu}.

\textit{Model Parallelism:} Each DNN convolution operation is independent of all other operations in the same layer. Model parallelism uses this intra-layer independence to split a DNN into disjoint subsets on multiple devices~\cite{strads}. %,dean. 
Model parallelism techniques usually have significant overhead in communicating activation maps~\cite{position}. 
These methods also suffer from the straggler effect when workloads are imperfectly balanced~\cite{wang2021sensAI, position}. Bhardwaj et al.~\cite{cmu} reduce the overhead of model parallelism, but their technique is applicable only when the number of devices is fixed and is known at training time.  
%When processing a frame, a large amount of intermediate results (e.g. feature- maps) need to be transferred among the GPUs that hold adjacent model partitions.

\textit{Pipeline Parallelism:} 
The DNN is divided into sets of consecutive layers, and each set is deployed on a collaborating device to improve throughput~\cite{pipeline, vcu}. Conventional DNNs (\eg VGG and ResNet) have been found to be ill-suited for pipeline parallelism because there is a large variance in resource requirements and communication costs across layers; this results in imbalanced workloads~\cite{pipeline}. Zhang et al.~\cite{pipeline} show that the inference time of the largest fully-connected layer of VGG-16 is $\sim15\times$ larger than the inference time of the smallest convolution layer on an edge-class device.
%When balancing workloads, the tradeoff between the number of collaborating devices and the number of layers per device increases either communication or computation costs~\cite{mednn}. 
To alleviate this issue at the cost of additional overhead, Hadidi et al.~\cite{pipeline2} combine pipeline parallelism with model parallelism to prevent bottlenecks.
%\jnote{Does it increase overhead by a marked amount?}

Table~\ref{tab:contributions} presents properties of existing techniques. The existing parallel techniques either balance loads or reduce communication costs~\cite{modnn, pipeline, pipeline2}.
We propose the first method to perform parallel inference of hierarchical DNNs. This method performs efficient pipeline parallelism by balancing loads and reducing overhead. This work enables the use of pipeline parallelism for improving the throughput in application contexts where multiple edge devices operate in close proximity (\eg airports, traffic intersections, construction sites, etc.).

%\anote{add context here}

%This tradeoff limits the performance gains possible with pipelining. When pipelining conventional DNNs, the number of collaborating devices needs to be known apriori to partition the DNN, thus limiting flexibility. Finally, partitioning DNNs with skip connections (e.g., ResNet, DenseNet, etc.) results in additional communication overhead.

% \subsection{}

\subsection{Our Contributions} 

As summarized by Table~\ref{tab:contributions}:
(1)~This is the first method to perform parallel hierarchical DNN inference to accelerate video stream processing on low-power embedded devices.
(2)~We develop a mathematical model to estimate the throughput gains with pipelined hierarchical DNNs in different application scenarios.
(3)~Using this model we present a novel technique that partitions the hierarchical DNN for maximizing throughput with pipeline parallelism. %The proposed method accounts for the hierarchy structure to reduce the parallel processing overhead.
(4)~We experimentally measure the factors that impact the throughput of the pipelined hierarchical DNNs. 
%\jnote{For pondering: Your partitioning scheme is static. Can you justify this?}

%\todo{grammar changes in this section}
%GKT: Reviewed this opening text and made a few minor grammatical improvements.

\section{Pipelining Hierarchical Neural Networks}

This section describes our pipeline parallelism scheme with hierarchical DNNs.
To create a hierarchical DNN inference pipeline, we first identify the factors that impact the processing time of pipeline-parallel hierarchical DNNs and create a model to estimate the throughput with our method (Section~\ref{sub:tradeoffs}).
We then use this model to find the hierarchy partition that maximizes the throughput (Section~\ref{sub:part}).

%will be lifted in our future work by accounting for the difference processing speeds of the devices when balancing loads and estimating throughput.

%\anote{add context here}

%To understand the working of pipeline-parallel hierarchical DNNs, consider the following example.
%Suppose a hierarchical DNN is partitioned to run on multiple devices as in Fig.~\ref{fig:pipeline}(b), the root DNN on device 1 processes frame 1. Once the root DNN makes an intermediate classification (\eg to DNN\textsubscript{2} on device 2), the frame gets passed to device 2. As device 2 processes frame 1, device 1 can begin processing frame 2. Frame 1 is next passed to device 4, frame 2 to device 4, and device 1 processes frame 3. In this example, before the output for frame 1 is generated, frame 2 and frame 3 are in the pipeline. Multiple frames are processed simultaneously to increase the FPS.
%\jnote{That was a lot of space to introduce this concept. Is it necessary? Can you slightly modify Figure 1 to show the frames there instead, and reduce the prose here? (I know we used to have a larger figure for this purpose, but I wonder if it can be squeezed into Figure 1 instead...)}

% \jnote{I added the Section refs into the numbered list in the previous paragraph. I suggest we cut this paragraph now.}
% The method to partition hierarchical DNNs to balance workloads across collaborating devices is described in Section~\ref{sub:part}. Section~\ref{sub:tradeoffs} analyzes different properties of hierarchical DNNs that impact the performance of the proposed distributed pipeline scheme.

\subsection{Throughput of Pipeline-Parallel Hierarchical DNNs}

\label{sub:tradeoffs}

%\textbf{TO COME: (a) EQUATION TO ESTIMATE SPEEDUP FOR DIFFERENT COMPUTE/COMMUNICATION TIMES. (b) FLOW CHART SHOWING HOW TO BUILD PIPELINE.}

%\jnote{GKT and I discussed potentially adding a citation for the ``style'' of the approach: classical methods to partition a computation graph. Not saying the results here are not novel (the details for H-DNN are new), but acknowledging the place where this contribution fits in the scholarly dialogue. Personally, I recommend 2 sentences after the initial (1) (2) (3) with 2-4 citations.}

%On a single device, the throughput of a hierarchical DNN is $(\#\textit{frames} \times \#\textit{DNNs} \times \textit{time to process DNN})$. 

In pipeline-parallel systems, the throughput depends on (1)~the number of pipeline stages, (2)~the time taken to process each stage, and (3)~the communication overhead.
The processing time for $F$ frames is given by the general equation:
  $P_{time} = \large{[} \normalsize{(}F + \textit{\small{\#}\normalsize{pipeline stages}} - 1\normalsize{)}\times \normalsize{(}\textit{\normalsize{max stage processing time}}\normalsize{)} \large{]} + \normalsize{\textit{communication time}}$~\cite{book1}.
The steady-state throughput for a video stream is given by $\frac{F}{P_{time}}$. In the proposed technique, the hierarchy structure (depth, number of edges, etc.) and the method used to partition the hierarchical DNN to run on collaborating devices impact the throughput. These factors impact the number of pipeline stages, the processing time, and the overhead.

\begin{table}[t]
\caption{Symbols reference. $^*$: Values are obtained after the hierarchical DNN has been partitioned.}
\begin{tabular}{cl}\toprule
           \textbf{Symbol} & \textbf{Definition} \\\midrule
$N$ & Number of collaborating devices \\
        $F$ & Number of frames \\ 
        $\Lambda$ & Avg. DNN processing time \\ 
        $\tau$ & Avg. communication time between devices\\
        $K$ & Maximum hierarchical DNN depth\\ 
        $H^*$ &  Avg. number of edge cuts from root to leaf  \\
        $M^*$ &  Avg. number of DNNs running sequentially on one device \\ \bottomrule
\end{tabular}
    \vspace{-0.15in}
    \label{tab:symbols}
\end{table}

We use seven parameters to model the throughput of the proposed pipelined hierarchical DNNs. These parameters are listed in Table~\ref{tab:symbols}. 
The average DNN processing time, $\Lambda$, depends on each DNN's processing time and rate of use.
DNNs may have different rates of use, dependent on two factors.
First, the hierarchy structure affects the rate of use; the root DNN is used most often because it processes every video frame, while leaf DNNs (\eg DNN\textsubscript{7} in Fig.~\ref{fig:pipeline}) process only a small subset of the frames.
Second, a DNN's rate of use is application-dependent; \eg in an airport, people are more common than cats, and so the DNNs responsible for processing people will be used more often than those for cats.
In the same way, the average communication time, $\tau$, depends on the amount of data transferred in each hierarchy edge and the rate at which the edges are used. Each hierarchy edge between a parent and child is used at the same rate as the child DNN (the edge is used only when the child is used).

In pipelined hierarchical DNNs, the hierarchy depth, $K$, determines the number of stages in the pipeline. When the pipeline is full, a DNN at every level of the hierarchy is processing a frame. 
%\todo[inline]{JD: This sentence is a little confusing. Perhaps ``When the pipeline is full, every DNN at every level of the hierarchy is processing a frame'', though that depends on whether you need to group DNNs together because they are deployed on a single node.} \todo[inline]{GKT: Agree with JD here.} 
$(F + K - 1) \times \Lambda$ is the DNN processing time for $F$ frames when $N = K$ (all stages can run in parallel). When $N < K$, only $N$ DNNs can run in parallel. The hierarchy partition algorithm assigns DNNs to the collaborating devices. If multiple DNNs along one root-leaf path are assigned to the same device, then the DNNs run sequentially on the device. In Fig.~\ref{fig:pipeline}(b), when Device 4 runs DNN\textsubscript{6}, DNN\textsubscript{3} must wait before it can process the next video frame.
If $M$ is the average number of DNNs that run sequentially on a single device, then the total DNN processing time can be approximated as $( F + N - 1)\times (M \times \Lambda)$. 
%\todo[inline]{The preceding equation is mildly confusing because it combines two terms from the general equation for $T_{pipe}$. Perhaps you can group $(M \times \Lambda)$ together so the relationship to $T_{stage}$ is clearer, or you can introduce an explicit $T_{stage}$ into your general equation and define it separately.}

After hierarchical DNNs are partitioned to run on collaborating devices, a hierarchy edge is considered to be \textit{cut} if it spans partitions (or devices). When an edge cut is encountered, the DNN activation map needs to be communicated between devices. Thus, for each frame $H \times \tau$ is the communication overhead. The total communication time is $F\times (H \times \tau)$.

Similar to $P_{time}$, the total time taken to process $F$ frames is  approximately $\Big ( ( F + N - 1)\times (M \times \Lambda) \Big ) +  \Big ( F\times (H \times \tau) \Big )$. The estimated throughput, $T_{th.}$, of our method is given in eqt.~(\ref{eqn:one}).

%where,  $( F + M - 1)\times \frac{K}{M} \times \Lambda$ is the time spent in DNN processing and $F\times H \times \tau$ is the communication overhead.

\begin{equation}
    \centering
    T_{th.} \approx \frac{F}{\Big ( ( F + N - 1)\times (M \times \Lambda) \Big ) +  \Big ( F\times (H \times \tau) \Big )}
    \label{eqn:one}
\end{equation}

This model estimates the throughput for different hierarchy structures, devices, and communication media. 
A hierarchical DNN partition method is required to assign DNNs to devices. The partition must find a tradeoff between the workload size ($M \times \Lambda$) and the overhead ($H \times \tau$) to maximize $T_{th.}$.

%Next, we describe how to partition the hierarchical DNN to increase $T_{th.}$. When partitioning the hierarchy, $F$, $K$, and $N$ are fixed. 

% \todo[inline]{It would be good to list assumptions (including homogeneity) under a dedicated \textbf{heading}. That puts everything in one place. Depending on the content length, you can list them before the model.}

\textbf{Model Assumptions}: This model operates under the assumption that there is no temporal relationship between frames. 
For example, eqt.~(\ref{eqn:one}) may not accurately estimate the throughput for a video where all frames containing cats appear first, followed by all the frames containing trucks, and so on. We embed this assumption into our model by using averaged values in $H$ and $M$.
This assumption does not sacrifice generality for edge applications because different objects may appear at any time; \eg over a day traffic cameras see cars, bikes, etc. Furthermore, our analysis only considers the situation when all devices have the same hardware. This assumption suits edge-contexts where homogeneous edge devices are deployed to simplify device management~\cite{manage}.

\begin{figure*}[t!]
        \centering
        \includegraphics[width = \textwidth]{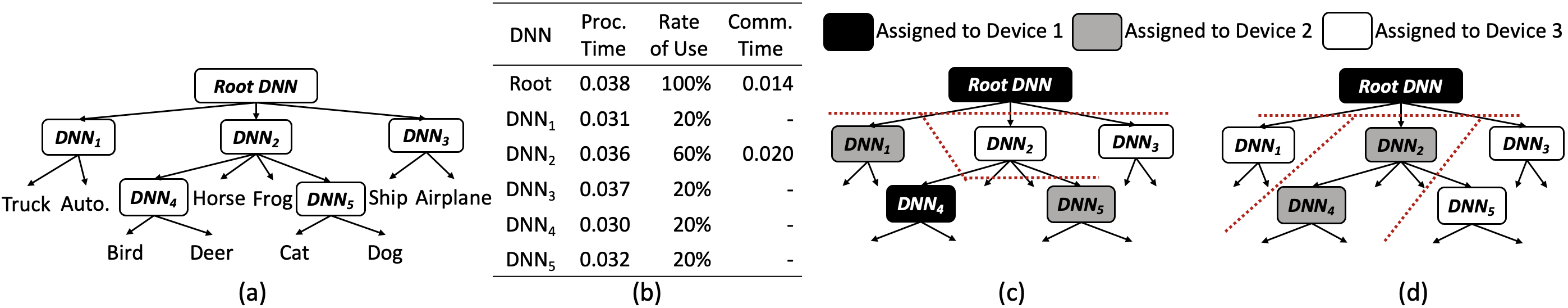}
        \caption{
        (a)~The hierarchical DNN structure constructed for the CIFAR-10 dataset using the methods described in Goel et al.~\cite{todaes}; along with (b)~the time taken to process a frame (in seconds), the estimated rate of use, and the communication time (in seconds).
        (c)~Hierarchy partition obtained when only balancing DNN processing times. (d)~Hierarchy partition obtained when balancing workloads and reducing communication. Dotted lines highlight the partitions. }
        \vspace{-0.15in}
        \label{fig:example}
\end{figure*}

%clarify if plausible, how to chnnage te model if neccessary. Which terms may change etc. etc.

%\todo[inline]{change WLOG statement}

%\jnote{Highlight the specific parameters that you consider fixed (\eg F, N), and the ones that you are tuning (H, $\Lambda$, $\tau$, $K$(?)).} \jnote{Equation 1 does not use symbol $K$ from Table II.}

\subsection{Partitioning Hierarchical DNNs for Pipeline-Parallelism}

\label{sub:part}

%\jnote{In the Background you did not mention this activation map. You should do so, because it hints about a reason to tackle the problems that you do.}

In this subsection, using examples in Fig.~\ref{fig:example}, we first show how the partition method impacts the throughput. We then describe our novel technique to find a hierarchy partition that maximizes the pipeline-parallel throughput.

\subsubsection{Impact of Hierarchical DNN Partitions on Throughput}

Hierarchical DNNs contain small DNNs in the form of a hierarchy.
To perform pipeline-parallel inference with hierarchical DNNs, first, the hierarchy must be partitioned. The hierarchy partition controls the values of $H$ and $M$ in eqt.~(\ref{eqn:one}).
Each partition is assigned to and run on a collaborating device. 
Hierarchies can be partitioned in different ways. %Fig.~\ref{fig:pipeline}(b) depicts only one way to partition a hierarchical DNN. 

To understand how the hierarchical DNN partition method impacts the DNN processing time and communication overhead, consider the example in Fig.~\ref{fig:example} when there are three collaborating devices. Fig.~\ref{fig:example} (a, b) depict the hierarchical DNN constructed for performing image recognition on images from the CIFAR-10 dataset along with the time taken to process images, the rate of use, the communication time for each DNN in the hierarchy. In this example, DNN\textsubscript{2} requires 0.036 seconds to process a frame and 0.020 seconds to communicate its activation map. Since the costs are comparable, finding hierarchy partitions that reduce the communication overhead is important for obtaining higher throughput. 
Without loss of generality, in this example, we assume that all the object categories in the dataset are equally probable to appear.
Thus, because there are 10 leaves in the hierarchy and 6 of those leaves are rooted at DNN\textsubscript{2}, the rate of use for DNN\textsubscript{2} is $\frac{6}{10}$. In other words, 60\% of the input frames will be processed by DNN\textsubscript{2}. If objects are not equally probable (\eg in an airport), the data must be sampled~\cite{data} to find the rate of use for each DNN before utilizing our approach. 
%\todo[inline]{GKT: Not a required paper change but potentially an opportunity to refine the conclusion here. If 6/10 are processed by one DNN and 4/10 by another, you should actually use this to increase the number of "servers" servicing the busier DNN. The ratio is 6:4, so this means some multiple of 3:2 devices, no? There's no harm to replicating devices (tree nodes) in this architecture as far as I can tell.}

%\todo[inline]{Is this assumption valid? Do you want to say something like "balanced samples" and cite some papers about unbalanced samples? Why is this assumption necessary? What happens if you do not have this assumption?}

If only the DNN processing times in Fig.~\ref{fig:example} are balanced, then $\langle$root, DNN\textsubscript{4}$\rangle$, $\langle$DNN\textsubscript{1}, DNN\textsubscript{3}$\rangle$, and $\langle$DNN\textsubscript{2}, DNN\textsubscript{5}$\rangle$ are assigned to the devices 1, 2, and 3, respectively, as shown in Fig.~\ref{fig:example}(c).
A hierarchy partition is balanced if the ratio of processing times on the most and least loaded  devices is minimized. Although the devices spend similar amounts of time running DNNs, the devices running the DNNs that are used more often will have larger workloads. To accurately account for workloads we must consider the hierarchy structure; \ie the rate at which DNNs are used. By scaling the DNN processing times by their rate of use, $\langle$root$\rangle$,  $\langle$DNN\textsubscript{3}$\rangle$, and $\langle$DNN\textsubscript{1}, DNN\textsubscript{2}, DNN\textsubscript{4}, DNN\textsubscript{5}$\rangle$ are assigned to  the devices.

%\todo[inline]{Will the temporal relationship also matter? In Figure 1, if DNN 4 is used many times first, then 5, then 6, then 7 (44444.... 55555.... ), will the result be different from alternating? (4567, 4567 ...)}

% 135
% 74
% 104.5

When only workloads are balanced, a single input frame's activation map may be communicated between multiple devices before the output is generated. In the previously described hierarchy partition, a frame that is processed by the DNNs along the path from the root to DNN\textsubscript{4} is communicated twice.
By choosing a partition that minimizes the edge cuts, the communication overhead can be reduced.
A balanced minimum cut graph partition is $\langle$root$\rangle$,  $\langle$DNN\textsubscript{3}, DNN\textsubscript{4}$\rangle$, and $\langle$DNN\textsubscript{1}, DNN\textsubscript{2}, DNN\textsubscript{5}$\rangle$. This hierarchy partition is depicted in Fig.~\ref{fig:example}(d). When performing image recognition with this partition, the expected processing time, $\sum ( \textit{DNN processing time} \times \textit{DNN rate of use})$, on the three devices are 0.038 seconds, 0.028 seconds, and 0.020 seconds, respectively. Similarly, the expected communication overhead of the hierarchy is given by $\sum ( \textit{Communication time} \times \textit{DNN rate of use}) = $ 0.018 seconds.

Having balanced workloads prevents bottlenecks, but may lead to larger communication overheads. Next, we discuss how to find hierarchy partitions that find a tradeoff between the workload balance and communication overhead to maximize the throughput, $T_{th.}$ in eqt.~(\ref{eqn:one}).

%\jnote{By now it is quite noticeable that I have had to hold a bunch of arithmetic in my head to follow along. I feel guilty about complaining, but also I am lazy. Can you make it easier for me to follow? See the suggestion in the caption of Figure 2.}
%This partition improves latency and throughput.

\begin{figure}[b!]
        \vspace{-0.15in}
        \centering
        \includegraphics[width = 0.47\textwidth]{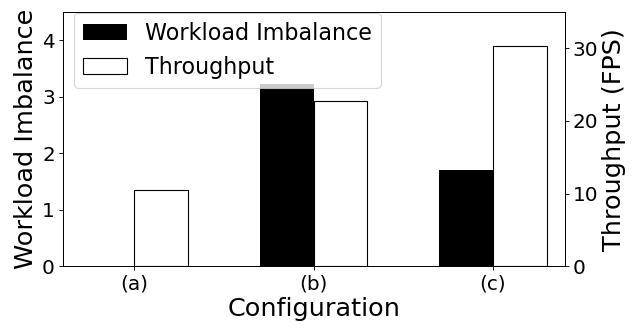}
        \caption{
            Impact of hierarchy partitions on throughput and workload imbalance. Workload imbalance is the ratio of the workloads on the most and least loaded devices.
            Configuration~(a) is depicted in Fig.~\ref{fig:example}~(a) for single-device inference. For three-device inference, configurations~(b) and (c) are depicted in Fig.~\ref{fig:example}~(c) and (d), respectively. Configuration~(c) achieves the highest throughput and the most balanced loads.
            %\todo[inline]{Can you use only one image (average time)? The other places use only per image, right? Please be consistent.}
        }
        \label{fig:compute_balance}
\end{figure}

%            \todo{You can shrink Figure 3 slightly for space, if you increase the font size of the x and y axis labels}

%Next, we discuss how to find the hierarchy partition that balances workloads and minimizes communication costs.

\subsubsection{Choosing Hierarchical DNN Partitions}

In this section, we use $D_i$ to represent the $i^{th}$ DNN of the hierarchy ($i = 0$ for root), and $E_j$ is a set representing the DNNs assigned to the $j^{th}$ collaborating edge device. $L(D_i)$ is the time taken (latency) by a device to process DNN $D_i$.
The communication time, $C_{i,j}$, is the time required to send an activation map from $D_i$ to $D_j$, when $D_i$ and $D_j$ are assigned to different devices.
$R(D_i)$ is the rate of use of DNN $D_i$. Because prior studies suggest that the largest hierarchical DNNs have only $\sim$20 DNNs~\cite{todaes} and our experiments consider a maximum of 4 collaborating devices, we use exhaustive search to find the hierarchy partition that maximizes the expected throughput. For larger hierarchies, heuristic algorithms like MeTis~\cite{metis} can be modified to find partitions without exhaustive search.

For every hierarchy partition, we first calculate the expected processing time on each device $E_j$ as $a_j = \sum L(D_i) \times R(D_i), \forall D_i \in E_j$. The value $a_j$ is the amount of time device $E_j$ takes to process an input scaled by the rate at which the DNNs are used. The largest expected processing time, $\text{max}(a_j)$, is the worst-case DNN processing workload on the devices. 
Having balanced workloads on devices minimizes $\text{max}(a_j)$.
We then obtain the parallel processing overhead of the partition in terms of the expected communication cost as $b = \sum C_{i,j} \times R(D_j), \forall \text{ DNNs }D_i, D_j$ that have a hierarchy edge that spans devices. Recall that the rate of use of an edge is the same as that of its child DNN.

To evaluate each hierarchy partition, we substitute $\Lambda \times M$ for $\text{max}(a_j)$, the largest expected processing time, and $H \times \tau$ for $b$, the expected communication cost, in eqt.~(\ref{eqn:one}). The estimated throughput is represented in eqt.~(\ref{eqn:two}). We select the hierarchy partition that maximizes the throughput, $T$. 

\begin{equation}
    T \approx \frac{F}{\Big ( ( F + N - 1)\times \text{max}(a_j) \Big ) +  \Big ( F\times b \Big )}
    \label{eqn:two}
\end{equation}

In light of this analysis, we can better understand Fig.~\ref{fig:example}. Fig.~\ref{fig:example}~(d) depicts the hierarchy partition obtained by maximizing $T$ for the CIFAR-10 dataset. When $\text{max}(a_j)$ is minimized the workloads are balanced across devices. When $b$ is minimized, the communication overhead is also minimized. Our proposed method finds a tradeoff between the workload balance and communication overhead to maximize the throughput, $T$.
Fig.~\ref{fig:compute_balance}~(c) shows this method processes $\sim$30 FPS with three devices collaborating over Ethernet, thus indicating efficient pipeline parallelism. Fig.~\ref{fig:example}~(c) shows another hierarchy partition that does not balance workloads or maximize $T$. The throughput obtained with this partition is 22~FPS, as seen in Fig.~\ref{fig:compute_balance}~(b).

\section{Experimental Results and Analysis}

\label{sec:expt}

This section experimentally evaluates the proposed method and existing techniques.
We vary the hierarchy structures, the edge devices, and the communication medium in our tests. The source code is available on Github~\cite{source}.

%The proposed approach yields between X\% and Y\% more energy-efficient inference than the nearest comparison point.

%The performance model (equation~(\ref{eqn:one})) predicts performance within \jnote{XX}\%.

% \iffalse
% This section shows that our implementation of the proposed hierarchy partitioning method results in significant video processing speedups.
% Experiments demonstrate that each device requires less memory and performs fewer operations, thus resulting in more energy-efficient inference. We also show the ability of the performance model, described in equation~(\ref{eqn:one}), to estimate the time taken by pipelined hierarchical DNNs to process data in different scenarios.
% %has competitive accuracy, and lower resource requirements than existing techniques. We also show that the method to construct the hierarchy using both visual and semantic similarities is important for the performance gains. 
% \fi

\subsection{Experimental Setup}
\textbf{Platform}:
We use up to four Raspberry Pi 4Bs in our experiments.
The devices communicate using gigabit Ethernet, underneath the ZeroMQ~\cite{zmq} message passing framework.
A NETGEAR Nighthawk AC5300 router is used for networking. The hierarchical DNNs open-sourced by Goel et al.~\cite{todaes} are used in our experiments.

\textbf{Metrics}:
We evaluate our proposed method based on five metrics:
  (a) Maximum memory required by a collaborating device for a video frame, measured using the \texttt{torchsummary} library;
  (b) Maximum number of DNN operations (FLOPs) performed by a device for a video frame, measured using the \texttt{thop} library;
  (c) Maximum energy consumed by a device, measured using a Yokogawa WT310E Power Meter;
  (d) Speed, as latency (inference time for one frame) and throughput (frames per second); (e) Speedup, as the ratio of the throughput obtained with parallel and single-device inference.
  
\textbf{Datasets Used}: We build and train hierarchical DNNs for three vision datasets: (1)~CIFAR-10~\cite{cifar}, (2)~SVHN~\cite{SVHN}, and (3)~The random subset of CALTECH-256~\cite{Caltech} used in previous works~\cite{FALCON, todaes}.
The CIFAR-10 and SVHN datasets contain small images (32$\times$32 pixels). Images in CALTECH-256 range from 200$\times$200 to 1024$\times$768 pixels and represent real-life images closely.
The Linux \texttt{ffmpeg} utility converts images from these datasets into varying-length videos for experiments.
For these datasets, the image contents, labels, and hierarchical DNN structures vary significantly~\cite{todaes}, allowing us to examine different types of workloads.

%\jnote{What is the property of interest? For example, do the datasets have different data ontologies, which results in different H-DNNs from [15]? To talk about this, we may need to introduce more detail into 2 or possibly treat it in 3.}
%For the proposed method, the worst-case memory requirement and FLOPs are reported: the sum of the model sizes/FLOPs of the DNNs along the longest path from the root to a leaf.

\subsection{Experiment 1 - Latency and Throughput}

We measure the effect of varying datasets on the proposed approach.
Performing pipeline-parallel inference of hierarchical DNNs increases throughput, but the communication overhead may lead to increases in latency. 
TABLE~\ref{tab:latency_fps} shows the impact of the input resolution and hierarchy structure on the latency and throughput of pipeline-parallel hierarchical DNNs. 
The hierarchical DNN for CALTECH-256 has a maximum depth of 5 and accepts inputs of 224$\times$224 pixels. SVHN and CIFAR-10 have hierarchical DNNs with depth 2 and 3, respectively.
With two devices for CIFAR-10, our method has 50 ms latency but achieves 20 FPS because of the inference pipeline. When using three devices, the latency increases to 64 ms and the throughput increases to 30.30 FPS. On one device, the throughput is only 10.55 FPS. Thus, the speedups with two and three devices are 1.90$\times$ and 2.87$\times$, respectively. Because the hierarchy for the SVHN dataset has a depth of 2, the speedup is smaller. Although the communication overhead with the CALTECH-256 hierarchy is higher due to the larger resolution, our method achieves a 2.48$\times$ speedup with 3 devices. This experiment indicates that the proposed method increases hierarchical DNN throughput for varying hierarchy structures and input resolutions.

% \begin{table}[]
% \centering
% \begin{tabular}{|c|rr|rr|rr|}
% \hline
% \multirow{2}{*}{Dataset} &
%   \multicolumn{2}{c|}{Latency (ms)} &
%   \multicolumn{2}{c|}{Throughput (FPS)} &
%   \multicolumn{2}{c|}{Speedup} \\ 
%  &
%   \multicolumn{1}{c}{N = 2} &
%   \multicolumn{1}{c|}{N = 3} &
%   \multicolumn{1}{c}{N = 2} &
%   \multicolumn{1}{c|}{N = 3} &
%   \multicolumn{1}{c}{N = 2} &
%   \multicolumn{1}{c|}{N = 3} \\ \hline
%  CIFAR-10 & 50 & 64 & 20.00 & 30.30  & 1.90$\times$ & 2.87$\times$ \\ 
%  SVHN & 43 & 50 & 41.32 & 58.82 &  1.61$\times$ & 2.29$\times$ \\ 
%  C-256 & 592 & 592 & 2.53 & 3.99 & 1.57$\times$ & 2.48$\times$ \\ \hline
% \end{tabular}
% \caption{Latency, throughput, and speedup obtained with the proposed pipeline-parallel hierarchical DNN method for different datasets. C-256: CALTECH-256.}
% \label{tab:latency_fps}
% \end{table}

\begin{table}[]
\caption{Latency, throughput, and speedup obtained with the proposed pipeline-parallel hierarchical DNN method for different datasets. C-256: CALTECH-256.}
\begin{tabular}{crrrrrr}\toprule
& \multicolumn{2}{c}{Latency (ms)} & \multicolumn{2}{c}{Throughput (FPS)} & \multicolumn{2}{c}{Speedup}
\\\cmidrule(lr){2-3}\cmidrule(lr){4-5}\cmidrule(lr){6-7}
           & N = 2  & N = 3    & N = 2  & N = 3 & N = 2 & N = 3\\\midrule
CIFAR-10    & 50 & 64 & 20.00 & 30.30  & 1.90$\times$ & 2.87$\times$ \\ 
SVHN & 43 & 50 & 41.32 & 58.82 &  1.61$\times$ & 2.29$\times$ \\ 
C-256 & 592 & 592 & 2.53 & 3.99 & 1.57$\times$ & 2.48$\times$\\\bottomrule
\end{tabular}

\label{tab:latency_fps}
\end{table}

\begin{table}[]
    \caption{Comparison of FLOPs ($\times 10^{6}$/frame), memory (MB/frame), energy (J/frame), and throughput (FPS) with different numbers of devices ($N$) for the CALTECH-256 dataset. Howard et al.~\cite{Mob} is a single-device method so ``-'' is used for $N > 1$. Blue font indicates the best result.}
    \centering
    \begin{tabular}{clrrrr}\toprule
               N &
  \multicolumn{1}{c}{Metric} &
  \multicolumn{1}{c}{\begin{tabular}[c]{@{}c@{}}Zhang\\ et al.~\cite{pipeline}\end{tabular}} &
  \multicolumn{1}{c}{\begin{tabular}[c]{@{}c@{}}Hadidi\\ et al.~\cite{pipeline2}\end{tabular}} &
  \multicolumn{1}{c}{\begin{tabular}[c]{@{}c@{}}Howard\\ et al.~\cite{Mob}\end{tabular}} &
  \multicolumn{1}{c}{\begin{tabular}[c]{@{}c@{}}Our Method\end{tabular}} \\\midrule
    \multirow{4}{*}{1} & \#Operations  & 15.51 & 4.11 &  \textcolor{blue}{0.58} & 1.38 \\
                  & Memory & 528.00 & 98.00 & 16.00 & \textcolor{blue}{6.20}   \\
                  & Energy & 27.72 & 14.76 &  13.86 & \textcolor{blue}{3.27}\\
                  & Throughput & 0.33 & 0.35 & 0.40 & \textcolor{blue}{1.62} \\\midrule
    \multirow{4}{*}{2} & \#Operations  & 9.37 & 2.97 &  - & \textcolor{blue}{0.92} \\
                  & Memory & 521.00 & 92.00 & - & \textcolor{blue}{5.00}   \\
                  & Energy & 19.72 & 8.56  & - & \textcolor{blue}{2.12} \\ 
                  & Throughput & 0.40 & 0.58  &  - & \textcolor{blue}{2.53} \\\midrule
    \multirow{4}{*}{3} & \#Operations  & 6.48 & 2.39  &  - & \textcolor{blue}{0.55} \\
                  & Memory & 521.00 & 65.00 & - & \textcolor{blue}{3.30}   \\
                  & Energy & 11.55 & 8.53 &  - & \textcolor{blue}{1.36} \\ 
                  & Throughput & 0.49 & 0.59 &  - & \textcolor{blue}{3.99} \\\midrule
    \multirow{4}{*}{4} & \#Operations  & 6.48  & 1.91  & - & \textcolor{blue}{0.55} \\
                  & Memory & 521.00  & 65.00  & - & \textcolor{blue}{2.60}   \\
                  & Energy & 10.90  & 8.46 &  - & \textcolor{blue}{1.04} \\ 
                  & Throughput & 0.52 & 0.60 &  - & \textcolor{blue}{5.20} \\\bottomrule
    \end{tabular}
    \label{tab:test}
    \vspace{-0.1in}
\end{table}

% \begin{table}[]
% \centering
% \begin{tabular}{|c|r|r|r|r|r|r|}
% \hline
% \multirow{2}{*}{Dataset} &
%   \multicolumn{2}{c|}{Latency (ms)} &
%   \multicolumn{2}{c|}{Throughput (FPS)} &
%   \multicolumn{2}{c|}{Speedup} \\ \cline{2-7} 
%  &
%   \multicolumn{1}{c|}{N = 2} &
%   \multicolumn{1}{c|}{N = 3} &
%   \multicolumn{1}{c|}{N = 2} &
%   \multicolumn{1}{c|}{N = 3} &
%   \multicolumn{1}{c|}{N = 2} &
%   \multicolumn{1}{c|}{N = 3} \\ \hline
%  CIFAR-10 & 50 & 64 & 20.00 & 30.30  & 1.90 & 2.87 \\ \hline
%  SVHN & 43 & 50 & 41.32 & 58.82 &  1.61 & 2.29 \\ \hline
%  C-256 & 592 & 592 & 2.53 & 3.99 & 1.57 & 2.48 \\ \hline
% \end{tabular}
% \caption{Latency, throughput, and speedup obtained with our pipeline-parallel hierarchical DNN method for different datasets. C-256: CALTECH-256.}
% \label{tab:latency_fps}
% \end{table}

\begin{figure*}[t!]
        \centering
        \includegraphics[width = \textwidth]{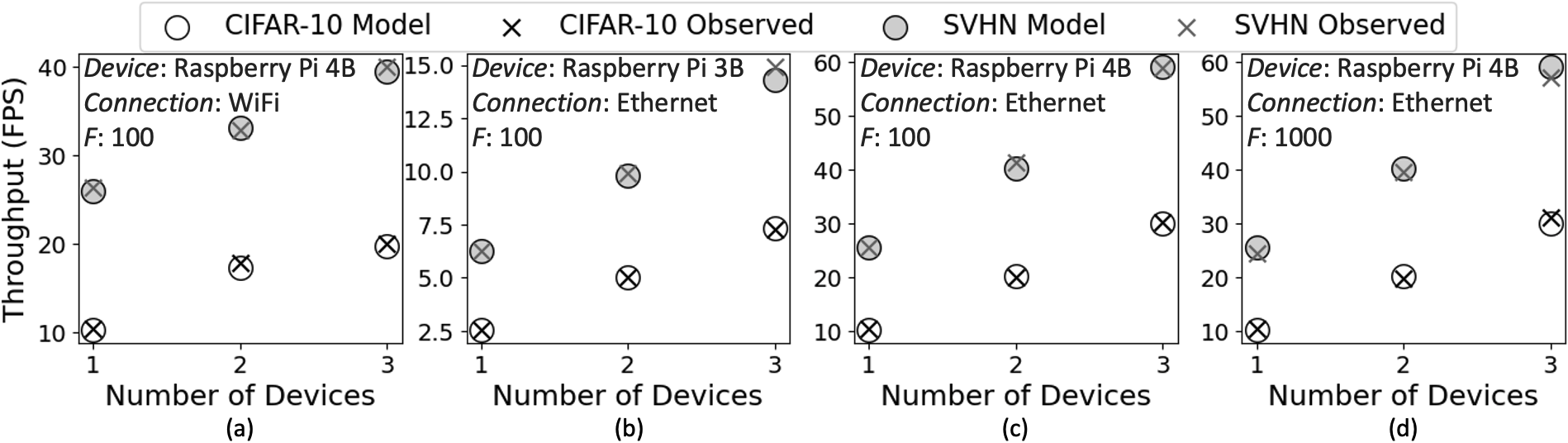}
        \caption{Evaluation of pipeline-parallel hierarchical DNN throughput model in eqt.~(\ref{eqn:one}). (a)~Raspberry Pi 4B connected with WiFi. (b)~Raspberry Pi 3B connected with Ethernet. Raspberry Pi 4B connected with Ethernet when the number of frames are: (c)~$F = 100$ and (d)~$F = 1000$. The observed results match the theoretical values closely in different application scenarios.
        }
        \vspace{-0.15in}
        \label{fig:res}
\end{figure*}

\subsection{Experiment 2 - Comparison with Existing Techniques}

We compare the proposed approach to the state-of-the-art, summarized in Table~\ref{tab:contributions}: single-device edge-friendly DNN inference~\cite{Mob}, single-device hierarchical DNN inference~\cite{todaes}, and parallel DNN inference (data~\cite{mednn, modnn}, pipeline~\cite{pipeline}, pipeline+model~\cite{pipeline2}).
%These experiments are conducted on the DNNs trained with the CALTECH-256 dataset. 
The results are tabulated in TABLE~\ref{tab:test}. Goel et al.~\cite{todaes} is the same as our method for $N =$1. As the number of devices increases from $N =$ 2 to 4, the memory required on a single device in Zhang et al.~\cite{pipeline} remains unchanged. This is because of the large variance in resource requirements across layers; large layers are not split onto multiple devices. Hadidi et al.~\cite{pipeline2} perform model parallelism to split large layers. However, because of the communication overhead, the time taken does not reduce significantly for $N > 2$. As $N$ increases, the proposed method finds hierarchy partitions that maximize the throughput resulting in significant reductions in processing time. With four devices, MoDNN~\cite{modnn} and MeDNN~\cite{mednn} achieve speedups of 2.03$\times$ and 2.43$\times$, respectively. In comparison, our method achieves a speedup of 3.21$\times$ with four devices, indicating more efficient parallelism. Results for MoDNN and MeDNN are not reported in TABLE~\ref{tab:test} because the data is not available.
We do not report accuracy because our method does not alter the accuracy of the existing hierarchical DNNs, it only increases efficiency.

%VRAI~\cite{VRAI} for vehicle re-identification, and Market-1501~\cite{Market} for person re-identification. VRAI contains 66,113 images of 6,302 different vehicles. Market-1501 contains over 32,000 images belonging to 1,501 different identities. This dataset also contains 500,000 distractor images for testing. Both datasets are divided into training and testing sets and are annotated with attributes~\cite{sem1}. 

% \subsection{Tradeoff between Latency and Throughput}

% Performing parallel pipeline inference of hierarchical DNNs increases throughput. However, the communication overhead leads to an increase in latency. To evaluate this tradeoff, we conduct experiments with different numbers of devices and different hierarchical DNN structures (constructed for different datasets). The partitions obtained for the different datasets is depicted in Fig.~\ref{fig:todo}. 

\subsection{Experiment 3 - Evaluation of Theoretical Model}

We evaluate the pipeline-parallel hierarchical DNN throughput model presented in eqt.~(\ref{eqn:one}). We use Raspberry Pi 3B and 4B boards in this experiment to vary values of $\Lambda$ (DNN processing times). To vary $\tau$ (communication time between devices), we use WiFi communication along with Ethernet. Finally, we also consider different hierarchy structures constructed for different datasets. Due to space constraints, we only present results for the CIFAR-10 and SVHN datasets, but we observe similar results with CALTECH-256 as well. For each run, first, a sample input is used to measure the values of $\Lambda$ and $\tau$. Then, the throughput is measured experimentally and compared with the theoretical value obtained with eqt.~(\ref{eqn:one}). Fig.~\ref{fig:res} shows that the observed experimental results match closely with the theoretical analysis for four different application scenarios.

\section{Conclusions}

In this paper, we present a novel method to perform pipeline-parallel inference of a hierarchical DNN for improving the processing throughput on low-power edge devices.
Our approach partitions the hierarchical DNN and deploys each partition on a collaborating edge device to allow the processing of multiple frames simultaneously. 
Existing pipeline-parallel DNN techniques partition conventional DNNs into sets of consecutive layers. These techniques are limited because the large variance in resource requirements and communication costs across layers creates bottlenecks in the pipeline.
Through this work, we present a method that partitions hierarchical DNNs to run on multiple devices with balanced loads and decreased communication costs.
We first mathematically model the throughput of pipeline-parallel hierarchical DNNs, and then find a hierarchy partition that maximizes the estimated throughput.
Our method can find appropriate hierarchy partitions automatically for varying hierarchical DNN structures, edge device specifications, and communication media.
Because of the hierarchy partition method, our pipeline-parallel hierarchical DNN achieves significant improvement in throughput with only a small increase in latency.
Our experiments confirm that the proposed inference strategy improves the deployability of computer vision on edge device networks, by decreasing the memory, energy, and number of operations on each device. 
%Our experiments also evaluate the pipeline-parallel hierarchical DNN throughput model. % is it s

%%%%%%%

%. %after review process.

\section*{Acknowledgement}

This project was supported in part by NSF CNS-1925713, NSF OAC-2107230,  NSF OAC-2104709, and NSF OAC-2107020. Any opinions, findings, and conclusions or recommendations expressed in this material are those of the authors and do not necessarily reflect the views of the sponsors.

\def\bibfont{\footnotesize}

\printbibliography
%\clearpage
%\tableofcontents
%\listoftodos

\end{document}